\documentclass[10pt,twocolumn,letterpaper]{article}

\usepackage{cvpr}              

\usepackage{graphicx}
\usepackage{amsmath}
\usepackage{amssymb}
\usepackage{booktabs}
\usepackage{multirow}
\usepackage{soul}

\usepackage[pagebackref,breaklinks,colorlinks]{hyperref}

\usepackage[capitalize]{cleveref}
\crefname{section}{Sec.}{Secs.}
\Crefname{section}{Section}{Sections}
\Crefname{table}{Table}{Tables}
\crefname{table}{Tab.}{Tabs.}
\newcommand{\para}[1]{\noindent\textbf{#1}}

\def\cvprPaperID{7841} 
\def\confName{CVPR}
\def\confYear{2022}

\usepackage{color}
\definecolor{green}{rgb}{0, 0.5, 0}
\definecolor{orange}{rgb}{0.8, 0.6, 0.2}
\definecolor{red}{rgb}{1.0, 0.0, 0.0}
\definecolor{teal}{rgb}{0.0, 0.4, 0.4}
\definecolor{purple}{rgb}{0.65,0,0.65}
\definecolor{saffron}{rgb}{0.95,0.75,0.2}
\definecolor{turquoise}{rgb}{0.0,0.5,0.5}
\definecolor{black}{rgb}{0.0, 0.0, 0.0}
\definecolor{gray}{rgb}{0.5, 0.5, 0.5}
\definecolor{blue}{rgb}{0.0, 0.0, 1}

\newcommand{\QM}[1]{{\color{black}#1}}

\newcommand{\rz}[1]{{\color{black}#1}}

\begin{document}

\title{UNIST: Unpaired Neural Implicit Shape Translation Network}

\author{Qimin Chen$^{1}$ \quad Johannes Merz$^{1}$ \quad Aditya Sanghi$^{2}$ \quad Hooman Shayani$^{2}$ \\ \quad Ali Mahdavi-Amiri$^{1}$ \quad Hao Zhang$^{1}$ \\
$^{1}$Simon Fraser University \quad \quad $^{2}$Autodesk AI Lab\\
}


\twocolumn[{%
\renewcommand\twocolumn[1][]{#1}%
\maketitle
\begin{center}
    \centering
    \includegraphics[width=0.99\textwidth]{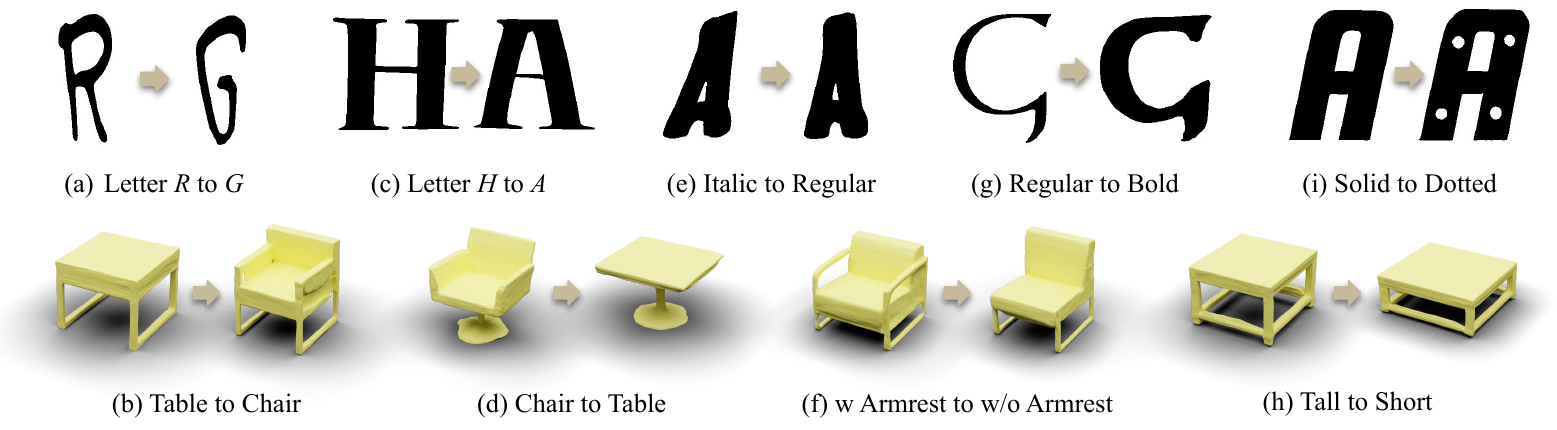} \vspace{-5pt}
    \captionof{figure}{We present UNIST, a model built on {\em neural implicit\/} representation that is able to learn both {\em style-preserving content alteration\/} (a-d) and {\em content-preserving style transfer\/} (e-i) between two {\em unpaired\/} domains of shapes, using the {\em same\/} network architecture.}
    \label{fig:teaser}
\end{center}%
}]


\begin{abstract}
We introduce UNIST, the first deep neural implicit model for general-purpose, unpaired shape-to-shape translation, in both 2D and 3D domains.
Our model is built on autoencoding implicit fields, rather than point clouds which represents the state of the art. Furthermore, our translation network is trained to perform the task over a {\em latent grid\/} representation which combines the merits of both latent-space processing and {\em position awareness\/}, to not only enable drastic shape transforms but also well preserve spatial features and fine local details for natural shape translations.
With the same network architecture and only dictated by the input domain pairs, our model can learn both style-preserving content alteration and content-preserving style transfer. We demonstrate the generality and quality of the translation results, and compare them to well-known baselines. Code is available at \small{\url{https://qiminchen.github.io/unist/}}.
\vspace{-5mm}
\end{abstract}


\section{Introduction}
\label{sec:intro}

Unpaired image-to-image translation has become one of the most extensively studied problems in computer 
vision since the advent of CycleGAN~\cite{zhu2017unpaired}, DualGAN~\cite{yi2017dualgan}, and UNIT~\cite{liu2017unsupervised} in 2017.
Somewhat surprisingly, there have been much fewer works on the same problem for shapes, i.e., {\em unpaired
shape-to-shape translation\/}.
To date, most image translation networks have been designed for style transfers that are localized, without large structural alterations. For shape translations, however, one may naturally expect more of the latter, e.g., to change 
the shape of a letter `R' to that of a `G', or a table to a chair; see Figures~\ref{fig:teaser}(a-d).

\begin{figure*}
  \includegraphics[width=0.99\linewidth]{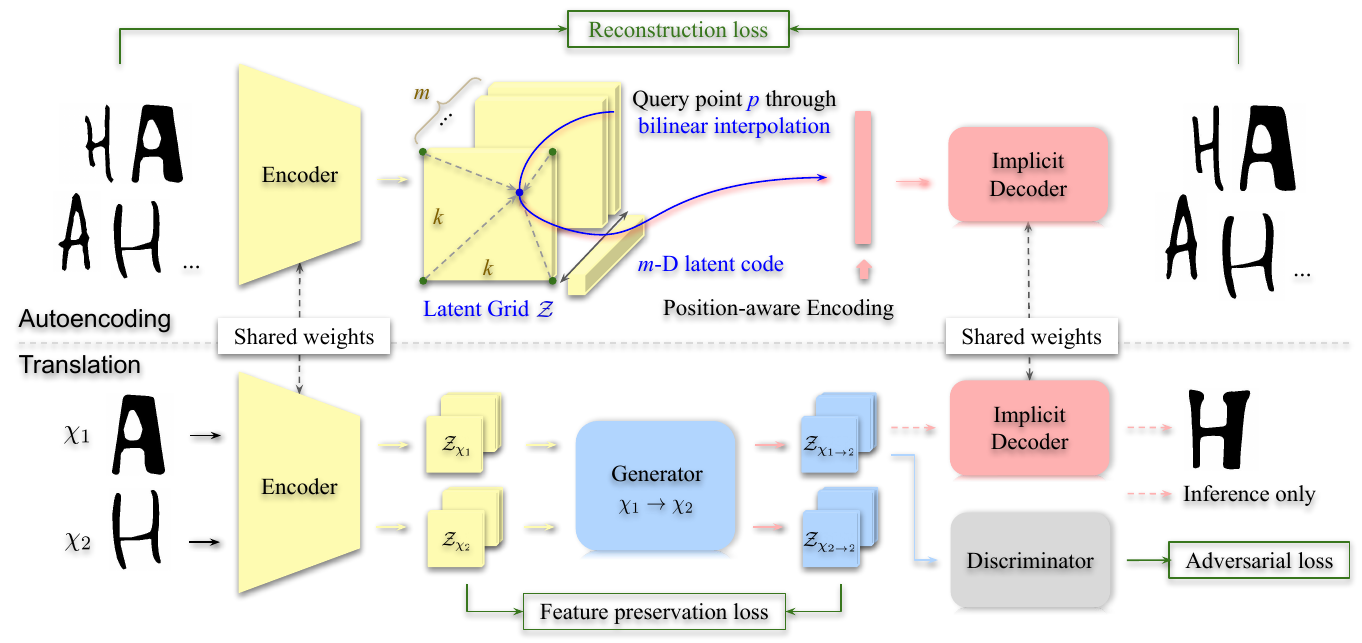}
  \caption{\rz{Overview of our framework for unpaired neural implicit shape-to-shape translation, which consists of two separately trained networks. The autoencoding network (top) learns to encode and decode binary voxel occupancies for shapes from both the source and target domains, where the encoder maps an input shape to a {\em latent grid\/} representation $\mathcal{Z}$. In the 2D case, $\mathcal{Z} \in \mathbb{R}^{k \times k \times m}$, where the $k \times k$ grid is obtained via spatial convolution over the $n \times n$ input image, and $m$ is the length of the latent code. The latent feature at any query point $p$ is obtained via {\em bilinear interpolation\/} over the latent codes stored in $\mathcal{Z}$. In the 3D case, the grid is three-dimensional and obtained via volumetric convolution and trilinear interpolation is performed to extract latent features for the decoder. The translation network (bottom) employs the pre-trained autoencoder network above to transform the translation problem into a latent space. In that space, a generator learns two tasks: 1) translating source-domain codes ($\mathcal{Z}_{\chi_{1}}$) into target-domain codes ($\mathcal{Z}_{\chi_{1\rightarrow2}}$); 2) preserving target-domain codes, from $\mathcal{Z}_{\chi_{2}}$ to $\mathcal{Z}_{\chi_{2\rightarrow2}}$.
$\mathcal{Z}_{\chi_{1\rightarrow2}}$ is passed to the pre-trained implicit decoder to obtain the final target shape resulting from the generator network.}}
  \label{fig:network}
\end{figure*}

Recently, Yin et al.~\cite{yin2019logan} proposed LOGAN, an unpaired shape translation network that can be trained to execute
both style and content, i.e., shape- and structure-level, transforms. However, their network was designed to operate on 
low-resolution point clouds (up to 2,048 points), which can severely limit the quality of the reconstructed and translated shapes,
especially in the 3D case.
In addition, the translation network is trained to operate on ``holistic'' latent codes which encode global information that is multi-scale but {\em not position-aware\/}. A consequence of such a lack of positional information in the encoding is losing control of spatial features, as well as local details, during shape translation. For example, when the translation is supposed to only italicize a letter shape, which is a pose change, the local details of a source shape (e.g., thickness/sharpness of certain tips of the shape) may be unexpectedly altered as well, as shown in the second row of Figure~\ref{fig:2d_reg_pos_retri} on the letter A translation.

In this paper, we present a method for unpaired shape-to-shape translation that is built on autoencoding {\em neural implicit
fields\/}~\cite{chen2018implicit_decoder,mescheder2019occupancy,park2019deepsdf}, rather than point clouds~\cite{yin2019logan}.
In recent years, advantages of learning continuous implicit functions over discrete representations
such as voxels, mesh patches, and point clouds have been demonstrated predominantly for reconstructive tasks 
including neural rendering, shape completion, and single-view 3D reconstruction. Our work shall show that the same
advantages of neural implicit models can be carried over to domain translation.

Furthermore, our translation network is trained to perform the task over a {\em latent grid\/} representation whose grid structure is {\em spatially correlated\/} with that of the input shapes, via convolution, while its remaining dimension encodes the latent features. Hence, our approach combines the merits of both latent-space processing and {\em position awareness\/}, with the former facilitating more drastic shape translations~\cite{liu2017unsupervised,yin2019logan} and the latter resulting in better preservation of spatial features and details~\cite{chibane2020implicit,peng2020convolutional,chen2021multiresolution} during the translation.

Our model, coined UNIST for unpaired neural implicit shape translation, consists of two separately trained networks, as illustrated in Figure~\ref{fig:network}. Given two {\em unpaired\/} domains of shapes, e.g., chairs and tables or various letters in different fonts (see Figure~\ref{fig:teaser} for several examples of domain pairs), the autoencoding network learns to encode and decode shapes (in the form of binary voxel occupancies) from {\em both\/} domains, using latent grids. The network training is self-supervised with the typical reconstruction loss.

The translation network is based on the LOGAN~\cite{yin2019logan} architecture which consists of a latent generator that is trained to perform two tasks: one is to translate the code of a source shape to that of a target shape under the adversarial setting, and the other is to turn the code of a target shape to itself based on a feature preservation loss. Differently from the original LOGAN, inputs to the UNIST generator are no longer the holistic and overcomplete latent codes; they are replaced by the latent grid features produced by the encoder of the pre-trained autoencoder network (top in Figure~\ref{fig:network}). The translation network is trained with the same set of losses as LOGAN, while the outputs from the generator would go through the pre-trained decoder (top in Figure~\ref{fig:network}) to produce the final shapes in the target domain. 

Our work represents the first deep implicit model for general-purpose, unpaired shape-to-shape translation. With the same network architecture and only dictated by the input domain pairs, our model can learn both style-preserving content alteration and content-preserving style transfer, as shown in Figure~\ref{fig:teaser}. We demonstrate the generality and quality of the translation results, and compare them to LOGAN~\cite{yin2019logan} and other baselines. We show that clear quality improvements on both shape reconstruction and translation are obtained merely by autoencoding implicit fields rather than point clouds. Further, adding position awareness~\cite{peng2020convolutional, chen2021multiresolution} through latent grids leads to more natural shape translation, with better preservation of spatial features and fine details.

\section{Related work}
\label{sec:related}

\para{Image translation.}
Image-to-image translation can be done under a paired or unpaired setting. An example of a paired translation is pix2pix \cite{isola2017image} that employs a conditional GAN with a reconstruction loss. Unpaired translation is more complex and usually requires additional loss functions such as cycle consistency \cite{zhu2017unpaired, yi2017dualgan}. While most methods \cite{zhu2017unpaired, yi2017dualgan} operate in image space, works such as \cite{liu2017unsupervised} have utilized a shared latent space to transfer more structural changes across two domains. However, such methods tend to produce less spatially aware local changes. Our work performs unpaired translation in latent space by using a latent grid with an implicit decoder, instead of a single holistic vector, to allow both structural and spatially aware local changes.  

\vspace{3pt}

\para{Shape translation.}
Paired shape-to-shape translation has been studied by P2P-Net \cite{yin2018p2p}, while Gao et al.~\cite{gao2018automatic} have explored unpaired deformation transfer. UNIST is inspired by LOGAN \cite{yin2019logan}, a general-purpose network for unpaired shape translation. 
However, our work differs from LOGAN in several significant ways. First, instead of point clouds, we employ implicit representations that can generate topology-varying shape translations. Second, we use position-aware latent grids rather than holistic latent codes that only encode global features. Finally, through extensive experiments, UNIST is shown to produce higher-quality reconstruction and more natural shape translation.

\vspace{3pt}

\para{Deep implicit functions.}
Implicit representations have gained immense popularity in the machine learning community and have recently been applied to 3D vision \cite{chen2018implicit_decoder, park2019deepsdf, mescheder2019occupancy}, images \cite{mildenhall2020nerf, tancik2020fourier, sitzmann2020implicit} and dynamic scenes \cite{li2020neural, xian2020space, park2020deformable}. 
Deep implicit representations can be broadly divided into three categories: global, local and hierarchical methods. Global methods use a single latent code for all location-based query points to decode objects and have been explored in works such as \cite{chen2018implicit_decoder, park2019deepsdf, mescheder2019occupancy, kleineberg2020adversarial, duan2020curriculum}. Whereas local implicit methods obtain a different latent code for each location-based query point by either tri-linearly interpolating on the encoder feature grid at a given point \cite{chibane2020implicit, peng2020convolutional} or just assigning a different latent code to each local part \cite{genova2020local, chabra2020deep, jiang2020local}. Recently, hierarchy-based methods have also been proposed \cite{chen2021multiresolution, liu2020neural} that incorporate level of detail to decode 3D shapes.  

Instead of global latent codes, we use latent grids as they have been successful in preserving details in autoencoding \cite{chibane2020implicit,peng2020convolutional}. Since our main goal is to perform \emph{translation} and translation at each query point is computationally expensive, we instead query latent grids for the decoder and perform the translation directly on latent grids.

\begin{figure}
     \centering
     \begin{subfigure}[b]{0.19\textwidth}
         \centering
         \includegraphics[width=\textwidth]{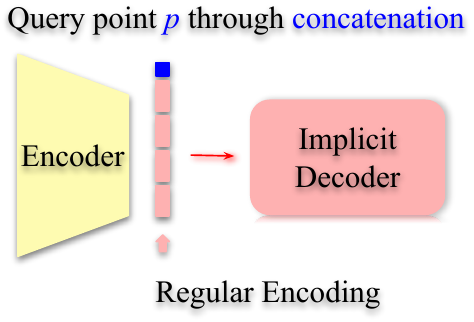}
         \caption{Regular encoding.}
         \label{fig:reg_encoding}
     \end{subfigure}
     \begin{subfigure}[b]{0.28\textwidth}
         \centering
         \includegraphics[width=\textwidth]{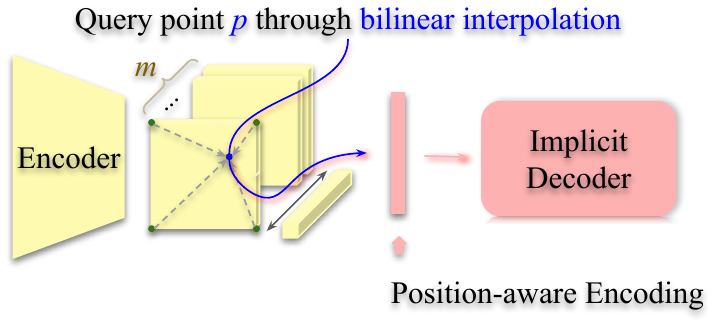}
         \caption{Position-aware encoding.}
         \label{fig:pos_encoding}
     \end{subfigure}
    \caption{Regular vs.~position-aware encoding. (a) The regular latent encoding is directly concatenated with point coordinate $p$ to predict inside/outside value. (b) The position-aware latent encoding is bilinear-interpolated from the latent grid centered at the point coordinate $p$ to predict inside/outside value.}
    \label{fig:reg_pos_encoding}
\end{figure}

\section{Methods}
\label{sec:methods}

In this section, we detail our design of UNIST, an implicit model for general-purpose, unpaired shape-to-shape translation in both 2D and 3D domains. As shown in Figure \ref{fig:network}, the network encodes input shapes from source and target domains with position-aware encoding and performs translation in a latent grid space, as guided by a set of losses to ensure domain translation as well as feature preservation.


\subsection{Neural implicit shapes}
\label{sec:neural_implicit_shapes}


Our network employs a neural implicit representation of shapes, by learning inside/outside or occupancy information~\cite{chen2018implicit_decoder,mescheder2019occupancy}. To reconstruct such shapes, the volumetric space around shapes is sampled. Denoting the input shape as $s$ and a query point as $p \in \mathbb{R}^{3}$, the implicit field $f$ is defined as:
    $f(p) = D(E(s), p) \rightarrow [0, 1],$
where query point $p$ along with the latent code of the input shape that is obtained from an encoder $E$ are passed to a decoder $D$ to predict 0/1 indicating inside/outside values. Then a mesh surface can be extracted 
via Marching Cubes \cite{lorensen1987marching}.

\subsection{Position-aware encoding}
\label{sec:Position_aware_encoding}

We adopt the idea of position-aware encoding from prior works, e.g.,~\cite{peng2020convolutional, chen2021multiresolution}, where the input shape is encoded into a latent grid $\mathcal{Z}$ that offers a latent code for each sample point in the space according to its position via linear interpolations. Instead of using a regular latent vector that is agnostic to spatial information, latent grids capture structural information about the input shape; see Figure~\ref{fig:reg_pos_encoding}.

More specifically, as illustrated in Figure \ref{fig:network} (top), given an input shape as a 2D binary image with size of $n\times n$, we first map it into a latent grid $\mathcal{Z}$ with size of $k \times k \times m$ using an encoder $E$ that consists of several 2D strided convolutions, and $m$ is the feature dimension. This latent grid $\mathcal{Z}$ encodes a compressed representation of the input shape with high-level spatial information. We then use bilinear interpolation to extract a latent code with size of $m\times1$ from the latent grid $\mathcal{Z}$ for query point $p$ according to its spatial position. This differs from \cite{mescheder2019occupancy, park2019deepsdf, chen2018implicit_decoder} where all the query points share the same latent vector, which results in networks paying more attention to global information and overlooking local details. A consequence of the absence of structural information in such an encoding is that when translating the latent space, a minor change in the latent vector could result in an unreasonable transform since all the information about the input shape is encoded in a single latent code. 

We choose IM-NET \cite{chen2018implicit_decoder} 
as our implicit decoder $D$ to decode the latent code interpolated from latent grid $\mathcal{Z}$ and predict inside/outside value for each point $p$. We optimize the position-aware encoding model for the reconstruction task using a weighted mean squared error between the ground truth SDF $\bar{s_{p}}$ and predicted SDF for query point $p$.
Let $\mathcal{S}$ be the training shapes and $\mathcal{P}$ be a set of points sampled on each shape, we then define the reconstruction loss as:
\begin{equation}
    \mathcal{L}_{recon} = \frac{1}{|\mathcal{S}|}\frac{1}{|\mathcal{P}|}\sum_{s\in\mathcal{S}}\sum_{p\in\mathcal{P}}\left | (D(\delta(E(s), p)) - \bar{s_{p}}) \cdot w_{p} \right |^{2}
\end{equation}
where $\delta(\cdot)$ denotes trilinear interpolation in the 3D case and bilinear interpolation in the 2D case, and $w_{p}$ denotes the weight assigned to point $p$ during sampling. More specifically, we set weights of points sampled near the boundary to 2 whereas weights of other points are set to 1. 
After training the autoencoder, we use its encoder and decoder as pre-trained networks for the translation task.

\subsection{Position-aware translation}
\label{sec:Position_aware_translation}

Unlike LOGAN, our generator takes the latent grid feature $\mathcal{Z}$ and translates it to match the target domain distribution while the discriminator learns the real distribution of the target domain over the \textit{entire} grid space; see Figure \ref{fig:network}. That is, given latent grids $\mathcal{Z}_{\chi_{1}}$ and $\mathcal{Z}_{\chi_{2}}$ with size ${k \times k \times m}$ from domains $\chi_{1}$ and $\chi_{2}$, the generator $G$ learns to translate $\mathcal{Z}_{\chi_{1}}$ to $\mathcal{Z}_{\chi_{1\rightarrow 2}}$ in the adversarial setting and $\mathcal{Z}_{\chi_{2}}$ to $\mathcal{Z}_{\chi_{2\rightarrow 2}}$ with feature preservation loss. Unlike LOGAN \cite{yin2019logan} whose discriminator works with a global latent vector and outputs a single scalar value, our discriminator $D$ takes in $\mathcal{Z}_{\chi_{1\rightarrow 2}}$ and outputs a scalar value for each latent vector in our $k \times k$ grid indicating whether features of each grid position are real or not. This results in more regulated translated feature vectors capable of carrying local and spatial information.

We optimize the translation network via losses over the latent grid space:
\begin{equation}
\label{eqn: overall_loss}
\mathcal{L}_{trans} = \mathcal{L}_{\chi_{1}\rightarrow\chi_{2}} + \mathcal{L}_{\chi_{2}\rightarrow\chi_{1}} + \gamma\mathcal{L}_{cycle}
\end{equation}

The loss function $\mathcal{L}_{\chi_{1}\rightarrow\chi_{2}}$ for translating domain $\chi_{1}\rightarrow\chi_{2}$ is defined as follows, $\mathcal{L}_{\chi_{2}\rightarrow\chi_{1}}$ can be easily defined by switching the domains:
\begin{equation}
\label{eqn: trans_loss}
    \mathcal{L}_{\chi_{1}\rightarrow\chi_{2}} = \mathcal{L}_{\chi_{1}\rightarrow\chi_{2}}^{\small{WGAN}} +  \alpha\mathcal{L}_{\chi_{1}\rightarrow\chi_{2}}^{GP} + \beta\mathcal{L}_{\chi_{1}\rightarrow\chi_{2}}^{FP}
\end{equation}
where $\mathcal{L}_{\chi_{1}\rightarrow\chi_{2}}^{\small{WGAN}}$ along with $\mathcal{L}_{\chi_{1}\rightarrow\chi_{2}}^{\small{GP}}$ are included in the usual WGAN loss with gradient penalty \cite{Gulrajani2017wgangp} and $\mathcal{L}_{\chi_{1}\rightarrow\chi_{2}}^{FP}$ is the feature preservation loss:
\begin{equation}
\mathcal{L}_{\chi_{1}\rightarrow\chi_{2}}^{\small{WGAN}} = \underset{z_{1}\sim\mathbb{P}(\mathcal{Z}_{\chi_{1}})}{\mathbb{E}}[D(G(z_{1}))] - \underset{z_{2}\sim\mathbb{P}(\mathcal{Z}_{\chi_{2}})}{\mathbb{E}}[D(z_{2})] \\
\end{equation}
\begin{equation}
\mathcal{L}_{\chi_{1}\rightarrow\chi_{2}}^{FP} = \underset{z_{2}\sim\mathbb{P}(\mathcal{Z}_{\chi_{2}})}{\mathbb{E}}[||G(z_{2})-z_{2}||_{1}]
\end{equation}

We enforce the network to naturally translate between two domains via cycle consistency loss defined as:
\begin{align}
\mathcal{L}_{cycle} & = \underset{z_{1}\sim\mathbb{P}(\mathcal{Z}_{\chi_{1}})}{\mathbb{E}}[||G_{2\rightarrow1}(G_{1\rightarrow2}(z_{1}))-z_{1}||_{1}] \nonumber \\ & + \underset{z_{2}\sim\mathbb{P}(\mathcal{Z}_{\chi_{2}})}{\mathbb{E}}[||G_{1\rightarrow2}(G_{2\rightarrow1}(z_{2}))-z_{2}||_{1}]
\end{align}
where $G_{1\rightarrow2}$ and $G_{2\rightarrow1}$ are used for $\chi_{1}\rightarrow\chi_{2}$ and  $\chi_{2}\rightarrow\chi_{1}$ translations, respectively.

\subsection{Implementation and training details}
\label{sec:impl_det}

In 2D experiments, we use $n=256$ pixels, $k=2$, and $m=64$.
For training our autoencoding network, we use a simple 2D conv-encoder where each layer downsamples images by half, and doubles the number of feature channels. For training the translation network, we use a generator with five 2D convolutional layers and a discriminator with four 2D convolutional layers. For autoencoding, we train all the 2D experiments for 800 epochs with batch size $24$, and use Adam optimizer and initial learning rate $0.00005$. We decay the learning rate by half after 400 epochs. 
For translation, we train the generators and discriminators for 1,200 epochs with batch size $128$, we again use Adam optimizer with initial learning rate $0.002$ and we halve the learning rate every 100 epochs until it reaches to $0.0005$. We empirically set $\alpha=10$, $\beta=20$ and $\gamma=20$ for Equation \eqref{eqn: overall_loss} and \eqref{eqn: trans_loss}.


We run all tests on a machine with two Nvidia GeForce GTX 1080 Ti GPUs. Training the autoencoder and translator networks takes 10 hours and 30 minutes, respectively, on 2D data. For 3D data, the times are 24 hours and 56 minutes, averaged over the object categories. Inference is fast: $~0.2$s per 2D shape; $~2.7$s and $~4.2$s for 3D outputs at $64^3$ and $256^3$ resolutions. More details on the network architecture and training can be found in the supplementary material.

\section{Experiments and results}
\label{sec:experiments_and_results}
\QM{
We first validate our network design by investigating the behavior of the regular encoding and position-aware encoding in Section \ref{sec:Ablation_study}. We then demonstrate the superior performance of UNIST on 2D shape translation in Section \ref{sec:Translation_on_2D_shapes} and 3D shape translation in Section \ref{sec:Translation_on_3D_shapes}.
}

\begin{figure*}[t!]
  \centering\includegraphics[width=0.97\linewidth]{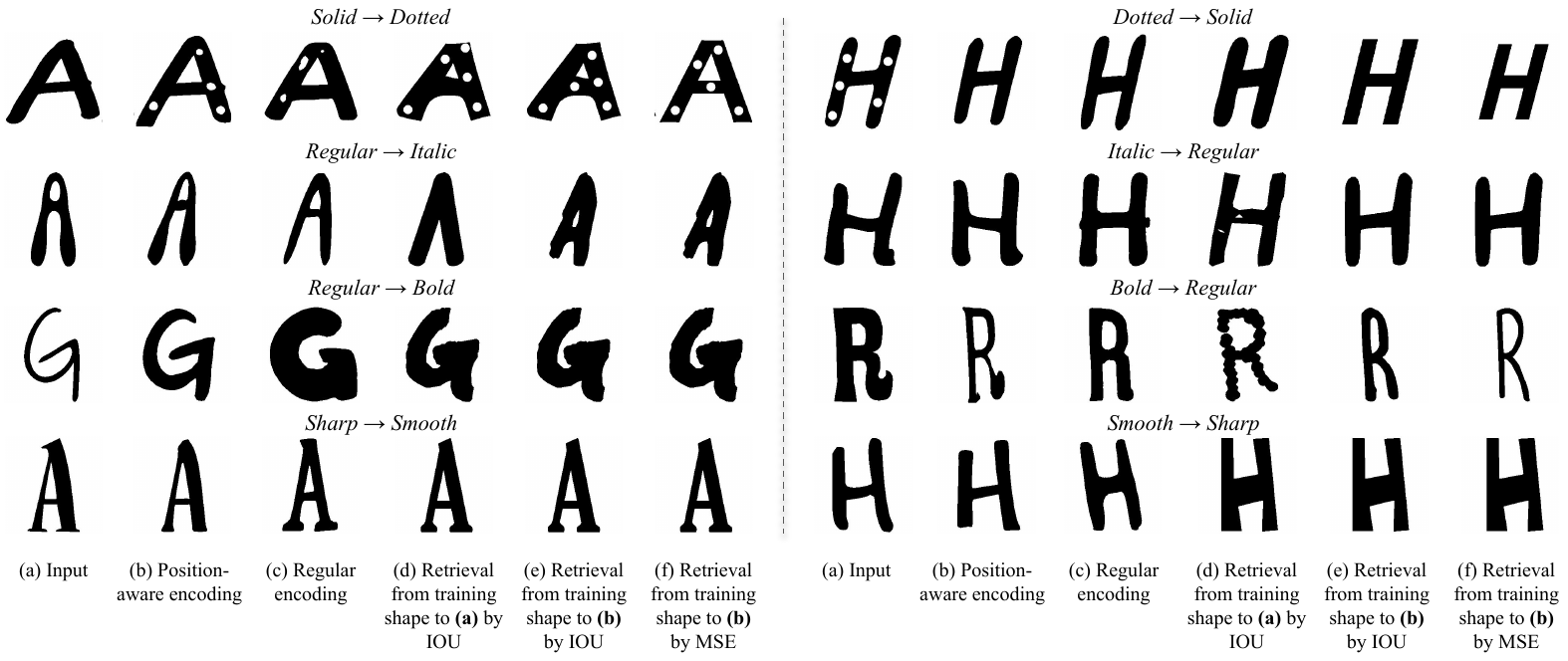}
  \caption{Qualitative comparison of position-aware encoding, regular encoding and retrieval results on different font shape datasets. First row (left): $Solid$$\rightarrow$$Dotted$, (right): $Dotted$$\rightarrow$$Solid$. Second row (left): $Regular$$\rightarrow$$Italic$, (right): $Italic$$\rightarrow$$Regular$. Third row (left): $Regular$$\rightarrow$$Bold$, (right): $Bold$$\rightarrow$$Regular$. Fourth row (left): $Sharp$$\rightarrow$$Smooth$, (right): $Smooth$$\rightarrow$$Sharp$. (a) Test input image. (b) Translation results by position-aware encoding. (c) Translation results by regular encoding. (d) Retrieved training shapes from the \textit{target} domain that are closest to test inputs (a) based on IOU measurement. Retrieved training shapes from the \textit{target} domain that are closest to position-aware encoding translation (b) based on (e) IOU measurement and (f) MSE measurement.}
  \label{fig:2d_reg_pos_retri}
\end{figure*}

\begin{figure*}[t!]
  \centering\includegraphics[width=0.99\linewidth]{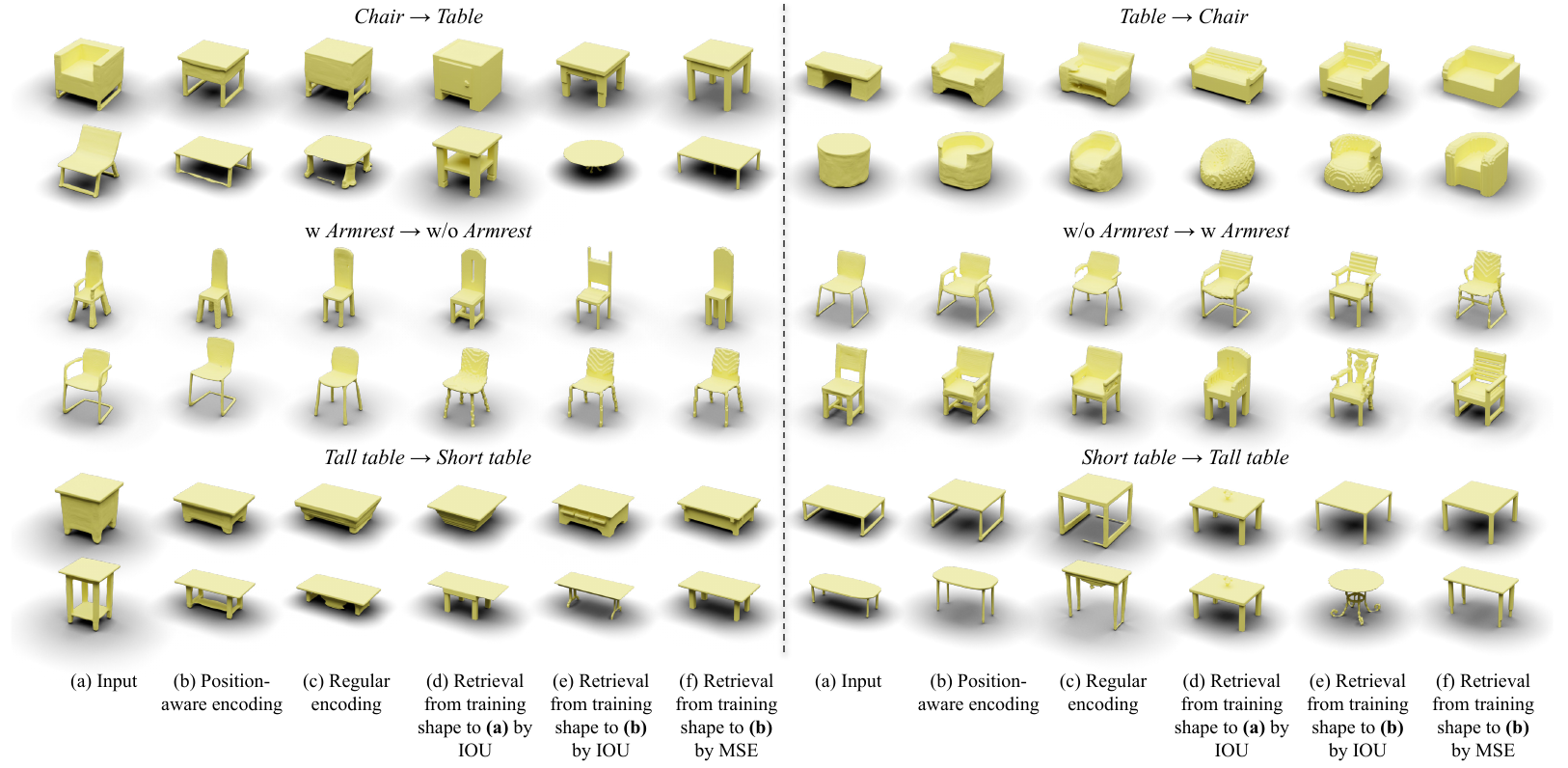}
  \caption{Qualitative comparison of position-aware encoding, regular encoding and retrieval results on different 3D shape datasets. Row 1-2 (left): $Chair$ $\rightarrow$ $Table$, (right): $Table$ $\rightarrow$ $Chair$. Row 3-4 (left): with $Armrest$ $\rightarrow$ without $Armrest$, (right): without $Armrest$ $\rightarrow$ with $Armrest$. Row 5-6 (left): $Tall \, table$ $\rightarrow$ $Short\,table$, (right): $Short\,table$ $\rightarrow$ $Tall\,table$. (a) Test input. (b) Translation results by position-aware encoding. (c) Translation results by regular encoding. (d) Retrieved training shapes from the \textit{target} domain that are closest to test inputs (a) based on IOU measurement. Retrieved training shapes from the \textit{target} domain that are closest to position-aware encoding translation (b) based on (e) IOU measurement and (f) MSE measurement.}
  \label{fig:3d_reg_pos_retri}
\end{figure*}

\QM{
\subsection{Ablation study}
\label{sec:Ablation_study}

\QM{
To verify the efficacy of position-aware encoding in translating shapes and preserving local details, we compare with our baseline, the regular encoding model, that is built on implicit autoencoding without position awareness.

\vspace{2pt}

\para{Regular encoding.} Our baseline model utilizes regular encoding (Figure \ref{fig:reg_encoding}) as opposed to position-aware encoding (Figure \ref{fig:pos_encoding}). The input shape is encoded into four sub-vectors with size of $m/4\times1$ from different convolutional layers in the encoder. These sub-vectors are concatenated to form an overcomplete latent code with size of $m\times1$ that is passed to the implicit decoder along with the coordinates of query point $p$ to predict its inside/outside value. 
}

\vspace{2pt}

\para{Regular vs.~position-aware encoding.} We compare translation results of the regular and position-aware encoding on four 2D cross-domain datasets: $Solid$ $\leftrightarrow$ $Dotted$, $Regular$ $\leftrightarrow$ $Italic$, $Regular$ $\leftrightarrow$ $Bold$ and $Sharp$ $\leftrightarrow$ $Smooth$. In these datasets, local geometric features, e.g., sharp/round corner and local curvature, are more prominent, resulting in content-preserving style transfer, which is suitable for validating the importance of position awareness in translation. We follow the same train and test split as in \cite{yin2019logan}. As shown in Figure \ref{fig:2d_reg_pos_retri} (b) and (c), it is evident that regular encoding manages to translate the input shapes into the target domain with fairly clean boundaries and is able to generate compact shapes. However, regular encoding is less capable of preserving the local geometric characteristics of each font (e.g., missing local curvature and irregular dots). We make use of three 3D cross-domain datasets: $Chair$ $\leftrightarrow$ $Table$, chair with $Armrest$ $\leftrightarrow$ without $Armrest$ and $Tall\,table$ $\leftrightarrow$ $Short\,table$ with same train and test split as \cite{yin2019logan} to further validate the position awareness. Similar observations are valid for 3D translation results illustrated in Figure \ref{fig:3d_reg_pos_retri} (b) and (c) where our network benefiting from position-aware encoding can produce results whose geometric and structural features have been better preserved. 

\vspace{2pt}

\para{Retrieval.} We show that UNIST is indeed aware of structural information embedded in the latent grid. That is, it generates a shape that is as similar to the input shape as possible and alters the shape to match the most distinctive features of the target domain. It is evident from Figures \ref{fig:2d_reg_pos_retri} (e-f), that translations with position-aware encoding are quite different from retrieved training shapes in the target domain. We also show retrieval results from the target domain that best match the test inputs and the translations with position-aware encoding on the 3D datasets in Figures \ref{fig:3d_reg_pos_retri} (d-f). One may notice that results in row 3, column (b) are similar to those in column (f), and the same in row 5. However, we can still observe differences in local details.
}

\QM{
\subsection{Translation on 2D shapes}
\label{sec:Translation_on_2D_shapes}

\begin{figure*}[t!]
  \centering\includegraphics[width=0.95\linewidth]{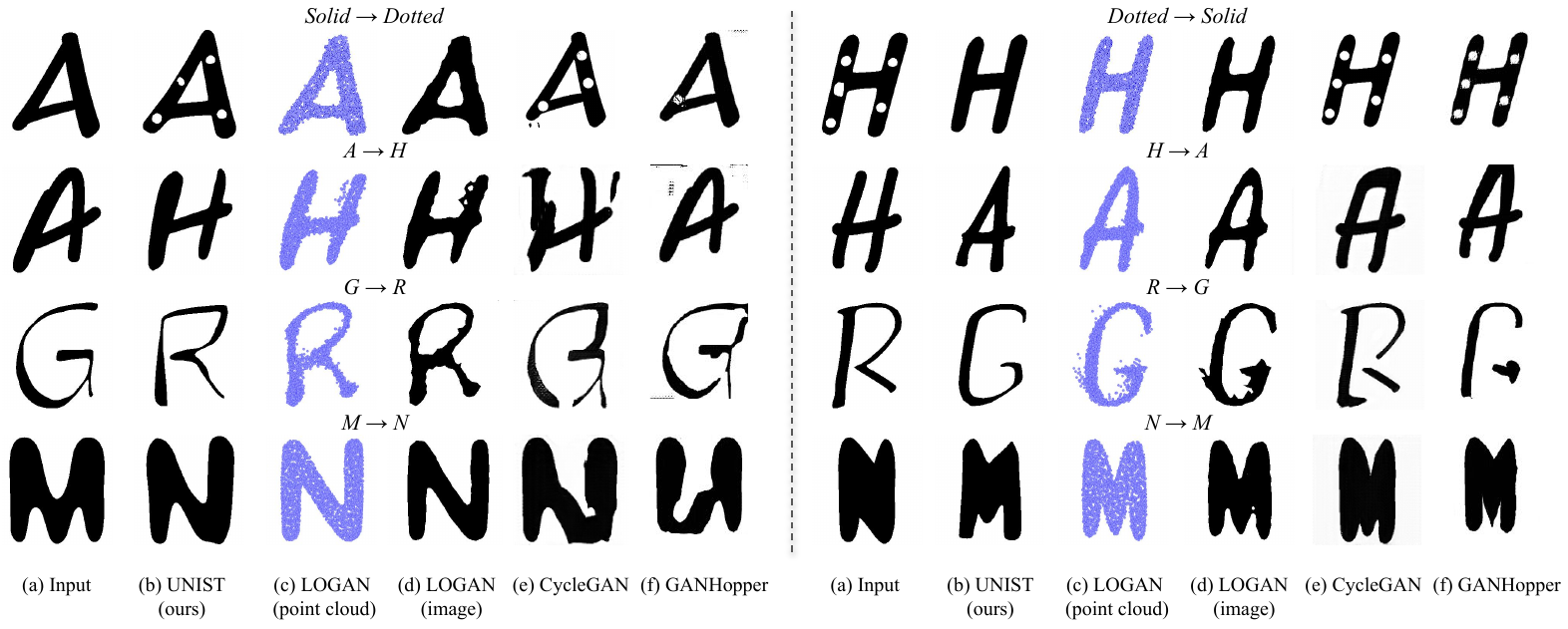}
  \caption{Comparison of translation results by UNIST (ours), LOGAN, CycleGAN and GANHopper on different 2D shapes. First row (left): $Solid$$\rightarrow$$Dotted$, (right): $Dotted$$\rightarrow$$Solid$. Second row (left): $A$$\rightarrow$$H$, (right): $H$$\rightarrow$$A$. Third row (left): $G$$\rightarrow$$R$, (right): $R$$\rightarrow$$G$. Fourth row (left): $M$$\rightarrow$$N$, (right): $N$$\rightarrow$$M$. (a) Test input. Translations by (b) UNIST, (c) LOGAN in point cloud representation, (d) LOGAN in image, (e) CycleGAN and (f) GANHopper.}
  \label{fig:2d_pos_logan_cyclegan_ganhopper}
\end{figure*}

For 2D shape translation, we compare UNIST to several unpaired cross-domain image-to-image translation networks: LOGAN \cite{yin2019logan}, CycleGAN \cite{zhu2017unpaired} and GANHopper \cite{lira2020ganhopper} on four datasets: $Solid$ $\leftrightarrow$ $Dotted$, $A$ $\leftrightarrow$ $H$, $G$ $\leftrightarrow$ $R$, and $M$ $\leftrightarrow$ $N$. Note that LOGAN uses point cloud representation for translation, hence, we fill the convex hull of the point clouds and convert to images for fair comparison. 

Figure \ref{fig:2d_pos_logan_cyclegan_ganhopper} shows the qualitative comparison of translation results. We observe that CycleGAN and GANHopper are less capable of learning both content-preserving style transfer ($Solid$ $\leftrightarrow$ $Dotted$) and style-preserving content transfer ($A$ $\leftrightarrow$ $H$, $G$ $\leftrightarrow$ $R$ and $M$ $\leftrightarrow$ $N$) while LOGAN manages to preserve and transfer features, it sometimes produces scattered point clouds, resulting in less compact shapes. Our proposed UNIST, on the other hand, is capable of producing shapes with significantly better visual quality, as it can reproduce small scale stylistic features as well as preserve topological features such as the dots in the dotted fonts.
}

\vspace{2pt}

\begin{figure}[t!]
  \centering\includegraphics[width=0.97\linewidth]{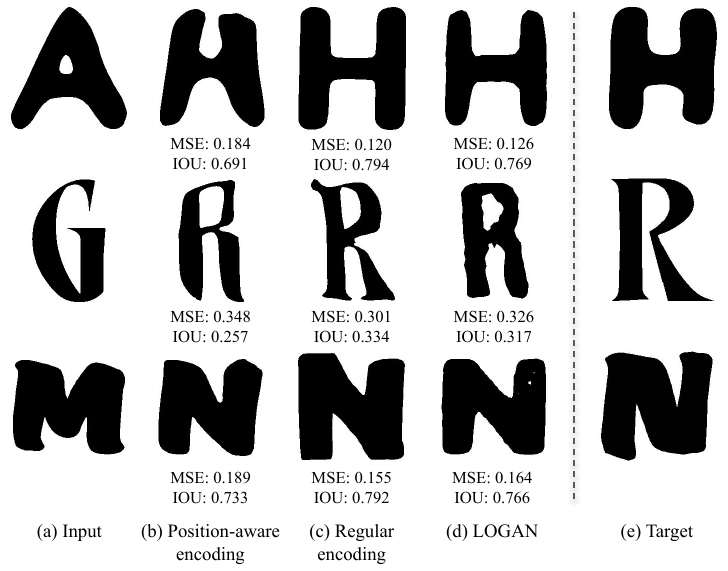}
  \caption{Visual comparisons between LOGAN and UNIST (regular vs.~position-aware encodings). 
 The target letters (e) are of the same font as the respective inputs (a). While position-aware encoding appears to produce more natural translations, with better spatial feature and style preservation, it is outperformed by the other two on MSE and IOU, measured against the targets.}
\vspace{-2mm}
  \label{fig:mse_iou_example}
\end{figure}

\rz{
\para{Quantitative comparison.}
We need ground-truth (GT) for a quantitative study. For $A$ $\leftrightarrow$ $H$, $G$ $\leftrightarrow$ $R$, and $M$ $\leftrightarrow$ $N$, a natural GT target would be letters from the same font family. Measured against that GT, using Mean Squared Error (MSE) and Intersection over Union (IoU), we show quantitative comparisons between various unpaired translation networks in Table \ref{tab:quantitative_with_correspondence}. As we can see, UNIST with regular encoding beats all competitors, including LOGAN, owing to the use of neural implicit shapes. On the other hand, the use of position-aware encoding underperforms against LOGAN, but still improves over other baselines. 

By a qualitative comparison shown in Figure \ref{fig:mse_iou_example}, we observe that position-aware encoding tends to preserve spatial and stylistic characteristics of the input as much as possible, only altering it in ways that are deemed most ``critical'' to reach the target domain. For the $A$ $\leftrightarrow$ $H$ example, only a small breaking was introduced at the top. Note that in Figure \ref{fig:mse_iou_example}, the regular encoding results may look closer to the targets. However, approaching the target is \textit{not} what the translators are trained to do as UNIST is fully \textit{unsupervised} and the translation results should be qualitatively judged by how well they preserve input features. Overall, we find position-aware encoding to produce more natural translation than regular encoding and LOGAN, e.g, see the $G$ $\leftrightarrow$ $R$ and $M$ $\leftrightarrow$ $N$ examples in Figure \ref{fig:mse_iou_example}. Its underperformance in Table \ref{tab:quantitative_with_correspondence} may be attributed to the strong feature preservation or the inadequacies of MSE and IoU as viable perceptual metrics as they only measure spatial distortions.


}

\begin{table}[t!]
    \begin{center}
    \resizebox{0.48\textwidth}{!}{\begin{tabular}{c|cc|cc|cc}
    \toprule
        & \multicolumn{2}{c|}{$A$ $\leftrightarrow$ $H$} & \multicolumn{2}{c|}{$G$ $\leftrightarrow$ $R$} & \multicolumn{2}{c}{$M$ $\leftrightarrow$ $N$} \\ \cmidrule(lr){2-3} \cmidrule(lr){4-5} \cmidrule(lr){6-7} & MSE $\downarrow$ & IoU $\uparrow$ & MSE $\downarrow$ & IoU $\uparrow$ & MSE $\downarrow$ & IoU $\uparrow$ \\
    \midrule
       CycleGAN & 0.246 & 0.385 & 0.229 & 0.412 & 0.266 & 0.383 \\
       UNIT & 0.253 & 0.376 & 0.264 & 0.377 & 0.295 & 0.348 \\
       MUNIT & 0.280 & 0.286 & 0.358 & 0.171 & 0.363 & 0.292 \\
       LOGAN & 0.195 & 0.490 & 0.213 & 0.472 & 0.207 & 0.506 \\
       GANHopper & 0.258 & 0.384 & 0.268 & 0.380 & 0.296 & 0.356 \\
    \cmidrule(lr){1-1} \cmidrule(lr){2-3} \cmidrule(lr){4-5} \cmidrule(lr){6-7}
       \multirow{2}{*}{\shortstack{Regular \\ encoding (ours)}} & \multirow{2}{*}{\textbf{0.184}} & \multirow{2}{*}{\textbf{0.507}} & \multirow{2}{*}{\textbf{0.197}} & \multirow{2}{*}{\textbf{0.493}} & \multirow{2}{*}{\textbf{0.201}} & \multirow{2}{*}{\textbf{0.508}} \\ & & & &\\
    \cmidrule(lr){1-1} \cmidrule(lr){2-3} \cmidrule(lr){4-5} \cmidrule(lr){6-7}
       \multirow{2}{*}{\shortstack{Position-aware \\ encoding (ours)}} & \multirow{2}{*}{0.215} & \multirow{2}{*}{0.441} & \multirow{2}{*}{\textbf{0.211}} & \multirow{2}{*}{0.452} & \multirow{2}{*}{0.233} & \multirow{2}{*}{0.433} \\& & & & \\
    \bottomrule
    \end{tabular}}
    \end{center}
    \vspace{-5mm}
    \caption{Quantitative comparisons between unpaired translation networks on $A$-$H$, $G$-$R$ and $M$-$N$ \rz{where {\em one possible\/} ground-truth (GT) target is available.} For each domain pair, Mean Squared Error (MSE) and Intersection over Union (IoU) are measured against {\em that\/} GT target letter and averaged over the translation in both directions (e.g. average over $A$$\rightarrow$$H$ and $H$$\rightarrow$$A$). $\downarrow$ means the lower the better and $\uparrow$ means the higher the better.}
    \vspace{-2mm}
    \label{tab:quantitative_with_correspondence}
\end{table}

\QM{
\subsection{Translation on 3D shapes}
\label{sec:Translation_on_3D_shapes}

\begin{figure*}[t!]
  \centering\includegraphics[width=0.99\linewidth]{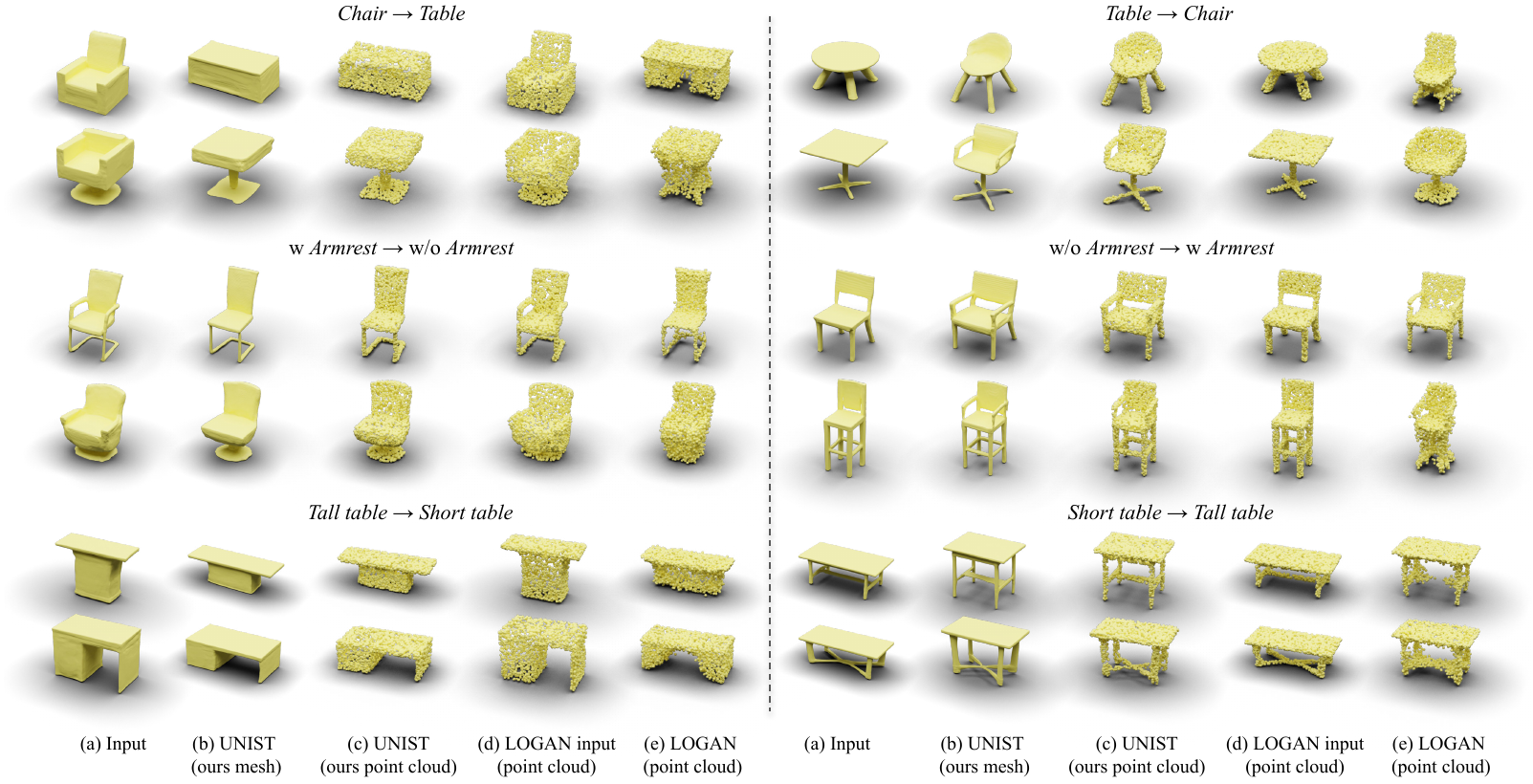}
  \caption{Comparison of translation results by UNIST (ours) and LOGAN on different 3D shapes. Row 1-2 (left): $Chair$ $\rightarrow$ $Table$ (right): $Table$ $\rightarrow$ $Chair$. Row 3-4 (left): w $Armrest$ $\rightarrow$ w/o $Armrest$ (right): w/o $Armrest$ $\rightarrow$ w $Armrest$. Row 5-6 (left): $Tall \, table$ $\rightarrow$ $Short\,table$, (right): $Short\,table$ $\rightarrow$ $Tall\,table$. (a) Test input mesh from voxel. (b) Translation by UNIST (ours) in mesh representation. (c) Translation by UNIST (ours) in point cloud representation. (d) LOGAN test input point cloud. (e) Translation by LOGAN.}
  \label{fig:3d_pos_logan}
\end{figure*}

We conduct 3D experiments on $Chair$ $\leftrightarrow$ $Table$, chair with $Armrest$ $\leftrightarrow$ without $Armrest$ and $Tall \, table$ $\leftrightarrow$ $Short\,table$ from ShapeNet \cite{chang2015_shapeNet}, and compare our method to LOGAN \cite{yin2019logan}, as it is the state-of-the-art unpaired shape-to-shape translation network. We use Marching Cubes \cite{lorensen1987marching} to obtain a mesh from the output, sampled at $256^3$ resolution. Note that we employ the sampling strategy from \cite{chen2018implicit_decoder} to obtain 2,048 points from the surfaces of the meshes to fairly compare with LOGAN, as it only produces point clouds at 2,048 resolution. Figure \ref{fig:3d_pos_logan} shows the qualitative results.
}

\rz{
Quantitative evaluations present challenges again, since shape translation is inherently a domain-specific task. What a \emph{correct} translation is can be highly varied, depending on the shape semantics from the two chosen domains and the nature of the translation itself. As a result, we provide different ways to evaluate 3D translations as follows.}

When translating between chairs with and without armrests (third and fourth rows of Figure~\ref{fig:3d_pos_logan}), our natural expectation is that the network should only add/remove the armrests while preserving the input. We treat this as the GT scenario and measure the quality of a translation using the {\em one-sided Chamfer Distance\/} (CD) from the armrest-less chair to its corresponding chair with armrests, regardless of the direction of the translation.
The numbers given in Table \ref{tab:w_arm_wo_arm_one_sided_cd} show that UNIST outperforms both our baseline and LOGAN on this metric, demonstrating that it better learns the essential difference between both domains.


In the case of $Table$ $\leftrightarrow$ $Chair$ translations, the network is tasked to not only modify the geometry of the input shape, but also change its semantics. Owing to this, it is hard to quantify a good result, motivating a user study to measure the quality of the translation. In our user study, we asked 72 participants via Amazon Mechanical Turk to rank the quality of translations performed by LOGAN \cite{yin2019logan} and UNIST, using regular or position-aware encoding. We report the study results in Table \ref{tab:user_study}, which show that the participants were most likely to choose UNIST with position-aware encoding as the best translation method over both directions. At the same time, position-aware UNIST was also least likely to be ranked as the worst of the three compared methods. Still, the gains are somewhat marginal, which is not entirely surprising given the ambiguity and subjectivity over how to judge what a good translation is. 

\begin{table}[t!]
    \begin{center}
    \resizebox{0.48\textwidth}{!}{\begin{tabular}{c|ccc|ccc}
    \toprule
         & \multicolumn{3}{c|}{Chair $\rightarrow$ Table} & \multicolumn{3}{c}{Table $\rightarrow$ Chair} \\ \cmidrule(lr){2-4} \cmidrule(lr){5-7} & 1st & 2nd & 3rd & 1st & 2nd & 3rd \\
    \midrule
       LOGAN & 30.55\% & 35.68\% & 33.25\% & 35.94\% & 30.47\% & 33.51\% \\ \cmidrule(lr){1-1} \cmidrule(lr){2-4} \cmidrule(lr){5-7}
       Regular & 31.34\% & 30.38\% & 37.67\% & 27.78\% & 36.37\% & 36.02\% \\ \cmidrule(lr){1-1} \cmidrule(lr){2-4} \cmidrule(lr){5-7}
       Position-aware & \textbf{38.11\%} & 33.94\% & \underline{29.08\%} & \textbf{36.28\%} & 33.16\% & \underline{30.47\%} \\
        \bottomrule
    \end{tabular}}
    \end{center}
    \vspace{-5mm}
    \caption{User study on $Chair$ $\leftrightarrow$ $Table$ via Amazon Mechanical Turk. Turkers were asked to rank translation results generated by different methods. $\%$ are relative to the total votes given per rank.} 
    \vspace{-2mm}
    \label{tab:user_study}
\end{table}

\begin{table}[t!]
    \begin{center}
    \resizebox{0.48\textwidth}{!}{\begin{tabular}{c|ccc|ccc}
    \toprule
         & \multicolumn{3}{c|}{w Arm $\rightarrow$ w/o Arm} & \multicolumn{3}{c}{w/o Arm $\rightarrow$ w Arm}\\
    \midrule
       {\small LOGAN} & \multicolumn{3}{c|}{{\small 0.0249}} & \multicolumn{3}{c}{{\small 0.0273}} \\ \cmidrule(lr){1-1} \cmidrule(lr){2-4} \cmidrule(lr){5-7}
       {\small Regular} & \multicolumn{3}{c|}{{\small 0.0255}} & \multicolumn{3}{c}{{\small 0.0267}} \\ \cmidrule(lr){1-1} \cmidrule(lr){2-4} \cmidrule(lr){5-7}
       {\small Position-aware} & \multicolumn{3}{c|}{\textbf{{\small 0.0234}}} & \multicolumn{3}{c}{\textbf{{\small 0.0235}}} \\
    \bottomrule
    \end{tabular}}
    \end{center}
    \vspace{-5mm}
    \caption{One-sided CD for translations between shapes with and without armrests to measure how well the common parts between both shapes are preserved. In the first result column, the one-sided distance was calculated as \textit{Output} $\rightarrow$ \textit{Input} and the second column represents \textit{Input} $\rightarrow$ \textit{Output}. In all calculations, we sample 2,048 points from the meshes to ensure a fair comparison to LOGAN.}
    \vspace{-2mm}
    \label{tab:w_arm_wo_arm_one_sided_cd}
\end{table}

\section{Conclusion, limitation, and future work}
\label{sec:future}

We show that the popular neural implicit representations are well suited to the task of unpaired shape-to-shape
translation, under the general framework of latent overcomplete GANs (LOGAN)~\cite{yin2019logan}. Improvements of UNIST over its point cloud counterpart are evident, especially when the translation or reconstruction involves finer details and topological changes. In addition, incorporating position-aware encoding into the design further strengthens the translation network in terms of feature preservation.

On the other hand, implicit functions are not as apt at representing geometric structures such as skeletons or profile curves, as point cloud LOGAN would. A more critical limitation however is related to {\em controllability\/}, or lack thereof. Hence the main path for future work is to explore few-shot learning and conditional generative modeling with UNIST to guide or constrain the translation network.

\vspace{3mm}
\noindent{\textbf{Acknowledgements}} We thank the annoymous reviewers for their comments, and Fenggen Yu, Wallace Lira, and Kangxue Yin for their discussions. This work was supported in part by NSERC (611370) and an Autodesk gift.

\clearpage
\newpage
{\small
\bibliographystyle{ieee_fullname}
\bibliography{egbib}
}

\end{document}